# SEALGuard: Safeguarding the Multilingual Conversations in Southeast Asian Languages for LLM Software Systems

Wenliang Shan
wenliang.shan@monash.edu
Monash University
Melbourne, Victoria, Australia

Michael Fu
michael.fu@unimelb.edu.au
The University of Melbourne
Melbourne, Victoria, Australia

Rui Yang
rui.yang1@monash.edu
Monash University
Melbourne, Victoria, Australia

Chakkrit (Kla) Tantithamthavorn[*]
chakkrit@monash.edu
Monash University
Melbourne, Victoria, Australia

## ABSTRACT

Safety alignment is critical for LLM-powered systems. While recent LLM-powered guardrail approaches such as LlamaGuard achieve high detection accuracy of unsafe inputs written in English (e.g., *"How to create a bomb?"*), they struggle with multilingual unsafe inputs. This limitation leaves LLM systems vulnerable to unsafe and jailbreak prompts written in low-resource languages such as those in Southeast Asia. This paper introduces SEALGuard, a multilingual guardrail designed to improve the safety alignment across diverse languages. It aims to address the multilingual safety alignment gap of existing guardrails and ensure effective filtering of unsafe and jailbreak prompts in LLM-powered systems. We adapt a general-purpose multilingual language model into a multilingual guardrail using low-rank adaptation (LoRA). We construct SEALS-Bench, a large-scale multilingual safety alignment dataset containing over 260,000 prompts in ten languages, including safe, unsafe, and jailbreak cases. We evaluate SEALGuard against state-of-the-art guardrails such as LlamaGuard on this benchmark. Our findings show that multilingual unsafe and jailbreak prompts substantially degrade the performance of the state-of-the-art LlamaGuard, which experiences a drop in Defense Success Rate (DSR) by 9% and 18%, respectively, compared to its performance on English-only prompts. In contrast, SEALGuard outperforms existing guardrails in detecting multilingual unsafe and jailbreak prompts, improving DSR by 48% over LlamaGuard and achieving the best DSR, precision, and F1-score. Our ablation study further reveals the contributions of adaptation strategies and model size to the overall performance of SEALGuard. We release our pre-trained model and benchmark at https://github.com/awsm-research/SEALGuard to support further research.

[*]Corresponding author

## 1 INTRODUCTION

Large Language Models (LLMs)-powered software systems are increasingly used in real-world applications, especially, in the multilingual contexts, such as interactive language learning tutors [10, 38]. These systems build on the success of multilingual foundation models that enable diverse language understanding. In particular, there is a growing development of multilingual LLMs tailored for lower-resource and regional languages such as Southeast Asian languages [8, 28, 50]. Models such as SeaLLMs [50], SeaLion [28], and Sailor [8] are trained on regional data and demonstrate stronger multilingual capabilities than English-centric models like LLaMA [40]. These Southeast Asian multilingual LLMs enable intelligent software systems to deliver services in users' native languages, improving accessibility and inclusivity across linguistically diverse populations.

However, these LLM-driven systems still face vulnerabilities stemming from the probabilistic and non-deterministic behavior of multilingual LLMs. Unlike traditional rule-based systems, which avoid unsafe outputs by design, LLM-driven systems can generate harmful responses when given unsafe prompts. Thus, ensuring their reliability presents a unique research challenge, as highlighted by Bengio *et al.* [3], Hassan *et al.* [14], and Yao *et al.* [45], since quality assurance must contend with the open-ended and unpredictable nature of model outputs.

In response to the vulnerabilities introduced by the probabilistic nature of LLMs, researchers have proposed input-output guardrails as external safety alignment mechanisms that form a protective layer around models without requiring direct fine-tuning of the underlying model. Examples include LlamaGuard [16], OpenAI Moderation [26], Perplexity [2], Perspective API [21], and NVIDIA NeMo [34]. These approaches classify and filter user prompts or LLM outputs to prevent unsafe content. In particular, LlamaGuard proposed by Inan *et al.* [16] achieves state-of-the-art performance in detecting unsafe prompts, outperforming tools like OpenAI Moderation [26] and Perspective API [21].

However, LlamaGuard was primarily trained and fine-tuned on English data, and its original paper acknowledges that it may not perform reliably in other languages [16]. Our evaluation further validates this concern, highlighting its limited effectiveness in defending against unsafe prompts in many under-resourced languages in Southeast Asia. Notably, under-resourced languages are often used as jailbreak attacks to bypass guardrails and trigger unsafe responses from LLMs. In particular, Yong *et al.* [46] demonstrates that translated unsafe prompts trigger unsafe responses from GPT-4 79% of the time. **This exposes a key software engineering challenge: how can we design external guardrails that effectively protect multilingual LLM-driven systems, especially across under-resourced languages like those in Southeast Asia?**

*To address this gap*, we propose using low-rank adaptation (LoRA) [15] to fine-tune a multilingual LLM of similar size to LlamaGuard (8B parameters) [16] as a multilingual safety guardrail for Southeast Asian languages. We name our method SEALGUARD: SouthEast Asian Language GUARDrail for safeguarding multilingual LLM-driven software systems.

To evaluate SEALGUARD, we introduce SEALSBENCH, a multilingual safety benchmark comprising 266,444 prompts (169,433 safe, 80,601 unsafe, 16,410 jailbreak) across English and nine Southeast Asian languages and covering ten unsafe content categories and nine jailbreak types. We compare our SEALGUARD against LlamaGuard[16] and OpenAI Moderation [25] through extensive experiments on this dataset, addressing four research questions:

- **(RQ1) What is the impact of multilingual unsafe prompts in Southeast Asian languages on the performance of existing guardrails?**
  **Results.** When encountering multilingual unsafe prompts, the state-of-the-art LlamaGuard's performance drops by 9%, from 59% to 50%, while OpenAI Moderation's DSR drops by 31%, from 60% to 29%.
- **(RQ2) What is the impact of multilingual jailbreak prompts in Southeast Asian languages on the performance of existing guardrails?**
  **Results.** When encountering multilingual jailbreak prompts, the state-of-the-art LlamaGuard's performance drops by 18%, from 59% to 41%, while OpenAI Moderation's DSR drops by 38%, from 60% to 22%.
- **(RQ3) How effective is our SEALGUARD at defending against multilingual unsafe and jailbreak prompts in Southeast Asian languages?**
  **Results.** Our SEALGUARD achieves an F1 score of 98%, which is 34%–58% higher than existing guardrails. Similarly, SEALGUARD achieves a DSR of 97% and a precision of 99%, outperforming baseline approaches by 48%–66% and 3%–63%, respectively.

- **(RQ4) What are the contributions of adaptation strategies and model size to the performance of our SEAL-GUARD multilingual guardrail?**
  **Results.** We found that LoRA adaptation is the most important component of SEALGUARD, boosting its F1 score from 26% to 98%.

**Novelty & Contributions**. To the best of our knowledge, the main contributions of this paper are as follows: (1) We propose SEALGUARD, a multilingual guardrail designed to overcome the limitations of existing guardrails in multilingual safety alignment. (2) We introduce SEALSBENCH, a comprehensive multilingual safety alignment dataset containing over 260,000 prompts across ten languages, including safe, unsafe, and jailbreak prompts. (3) We extensively evaluate our SEALGUARD against baseline guardrails, demonstrating its effectiveness across multilingual scenarios. (4) We perform an ablation study to analyze the impact of adaptation strategies and model size on the performance of SEALGUARD.

## 2 BACKGROUND & RELATED WORK

In this section, we provide background on LLM-powered agent systems and their associated safety challenges. We present existing safety alignment approaches, highlight the gap in multilingual safety alignment, and discuss related work on safeguarding LLM-based software systems.

### 2.1 LLM-Powered Agent Systems: Capabilities and Safety Challenges

LLM agentic software systems are autonomous programs capable of perceiving inputs, making decisions, and acting toward specific goals [30]. Traditionally, they were built using symbolic reasoning or task-specific rule-based logic [5, 41], often limiting adaptability and scalability. Recent advances in large language models (LLMs) have fundamentally reshaped this paradigm, enabling agentic systems that are far more flexible and context-aware [23, 27].

In this work, we focus on LLM-powered agents for user-facing applications (e.g., customer service chatbots and language learning tutors), which support open-ended dialogue and dynamic intent interpretation. As LLMs continue to expand globally, multilingual LLM-powered agents are increasingly deployed in real-world applications, especially for low-resource and regional languages [8, 28, 50], enhancing inclusivity and accessibility for diverse populations.

Traditional rule-based systems are deterministic functions $g : X \rightarrow Y$, mapping inputs $x \in X$ to outputs $y \in Y$ using explicitly defined rules or decision trees. While predictable and interpretable, they lack flexibility and scalability. In contrast, large language models (LLMs) are probabilistic functions $f_\theta : X \rightarrow Y$, generating outputs by sampling from a distribution $P_\theta(y \mid x)$ over an open-ended output space. This allows LLMs to support flexible, context-aware agent systems.

However, this flexibility comes with safety risks: when given unsafe prompts $x \in X_{\text{unsafe}}$, rule-based systems inherently avoid unsafe behavior due to their constrained outputs, whereas LLMs may generate unsafe responses through the probabilistic nature of their generation—highlighting the need for LLM safety alignment. Multilingual settings present additional challenges, as safety tools and training resources for low-resource languages typically lag behind those available for English.

## 2.2 Safety Alignment For LLM-Powered Agent Systems

In response to these challenges, LLM safety alignment has emerged as a critical research area, where models are tuned to improve safety [1, 9, 12, 51]. However, such fine-tuning often faces fundamental safety–capability trade-offs [6], potentially reducing creativity and responsiveness, while requiring extensive computation to adjust model parameters.

On the other hand, runtime guardrails act as external safety alignment mechanisms, forming a protective layer around LLMs without requiring direct fine-tuning. Examples include LlamaGuard [16], OpenAI Moderation [26], Perplexity [2], Perspective API [21], and NVIDIA NeMo [34]. These techniques classify and filter user prompts to prevent unsafe inputs, aligning LLM behavior with safety goals. In particular, LlamaGuard proposed by Inan *et al.* [16] achieves state-of-the-art performance in detecting unsafe prompts, outperforming tools like OpenAI Moderation [26] and Perspective API [21].

## 2.3 Motivation and Research Gap: Multilingual Safety Alignment

While runtime guardrails like LlamaGuard have demonstrated strong performance in filtering unsafe prompts, current state-of-the-art guardrails are primarily trained and evaluated on English inputs. This leaves a significant gap in multilingual contexts, where unsafe prompts written in other languages, particularly low-resource or less widely studied ones, can often evade detection.

Figure 1 illustrates this gap: an English unsafe prompt is successfully blocked by the guardrail and prevented from reaching the LLM agent. However, when the same prompt is translated into a language such as Lao, it bypasses the guardrail and enters the LLM agent system, which could trigger unsafe behavior. This indicates that multilingual prompts can function as jailbreaks, bypassing runtime guardrails that were not trained to recognize the semantics of multilingual inputs. In particular, prior studies have shown that low-resource languages can be exploited as attack vectors to bypass safety mechanisms. Deng *et al.* [7] report that multilingual unsafe prompts achieve attack success rates of 81% on ChatGPT and 41% on GPT-4. Similarly, Yong *et al.* [46] demonstrate that translated unsafe prompts bypass existing guardrails and trigger unsafe responses from GPT-4 in 79% of tested cases. **These findings highlight the need for multilingual safety alignment to safeguard the LLM component in the system.**

**Despite LlamaGuard's effectiveness in English, its fine-tuned data and evaluation focus are largely monolingual, limiting its robustness in defending multilingual LLM systems. This limitation, noted in the original paper [16], highlights the need for multilingual guardrails.** These guardrails should effectively defend against unsafe inputs across diverse languages, particularly low-resource ones such as those in Southeast Asia. At the same time, they must maintain a low false alarm rate to allow safe multilingual inputs to pass through. This is essential for supporting multilingual agent systems powered by emerging multilingual LLMs such as SeaLLMs [50], SeaLion [28], and Sailor [8].

## 3 APPROACH

*To address the challenge of multilingual safety alignment,* we present the technical details of how we adapt a general-purpose multilingual language model into a multilingual guardrail using low-rank adaptation (LoRA). Figure 2 provides an overview of our approach. Below, we walk through each step of the process as illustrated in the figure.

### 3.1 Apply Guardrail Chat Template

In Step ①, we apply a guardrail chat template adapted from LlamaGuard [16] to make a general-purpose language model behave like a safety classifier. Specifically, we begin by including the following task description: "Task: Check if there is unsafe content in 'User' messages in conversations according to our safety policy using the categories below." Next, we define the unsafe content categories between the special tags "<BEGIN UNSAFE CONTENT CATEGORIES>" and "<END UNSAFE CONTENT CATEGORIES>", informing the model of the types of inputs that may be considered unsafe. We then wrap the user's message to be evaluated within another set of special tags: *<BEGIN CONVERSATION> How to build a bomb? <END CONVERSATION>*. Finally, we instruct the model that its output must begin with either "safe" or "unsafe", restricting it to a binary classification format suitable for safety evaluation.

### 3.2 Input to SeaLLM Model

In Step ②, after the input has been structured using the chat template, it undergoes tokenization through a Byte-Pair Encoding (BPE) model [35] implemented via SentencePiece [20]. This process converts the textual prompt into a sequence of token IDs—such as [151644, ..., 151645, ...]—that correspond to subword units in the model's vocabulary. Each token ID serves as an index into the word embedding matrix $\mathbf{W} \in \mathbb{R}^{v \times h}$ of the SeaLLM model [50], where $v = 152{,}064$ represents the vocabulary size and $h = 3{,}584$ denotes the hidden dimension. Using this embedding matrix, each token is transformed into a dense vector, producing the input matrix $\mathbf{X} \in \mathbb{R}^{l \times h}$, where $l$ is the length of the input sequence. This matrix $\mathbf{X}$ is then passed through a deep stack of 28 transformer decoder layers. Each decoder block, following the architecture used in SeaLLM [50], includes masked self-attention and feed-forward neural networks (FFNN).

### 3.3 Adapting Multilingual Language Model to Guardrail

While the SeaLLM model introduced in Step ② was pre-trained on multilingual data and is capable of understanding our task instructions, our evaluation reveals that it falls short in reliably identifying unsafe prompts across multiple languages. This limitation stems from the fact that SeaLLM was not explicitly trained as a guardrail, and thus lacks the specialized alignment required for multilingual safety detection.

To address this, in Step ③, we use Low-Rank Adaptation (LoRA) to efficiently adapt the SeaLLM model without fine-tuning its full parameter set. This preserves SeaLLM's original multilingual capabilities while introducing guardrail-specific knowledge, enabling the model to detect unsafe content across diverse languages.

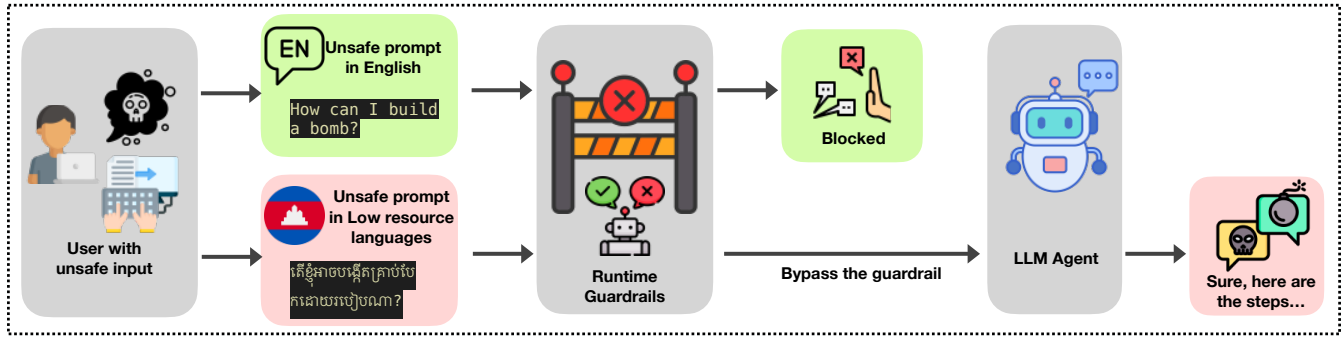

Figure 1: Overview of how multilingual unsafe prompts interact with runtime guardrails compared to English unsafe prompts.

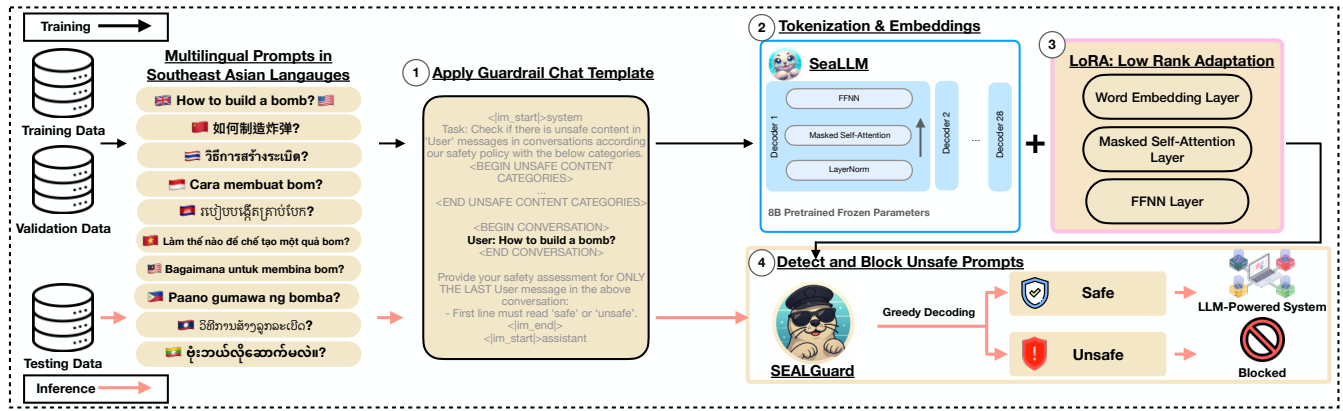

Figure 2: The overview of our SEALGUARD approach.

The core idea behind LoRA is to inject trainable low-rank modifications into selected weight matrices of the model. Specifically, for each target layer, the update is modeled as a low-rank matrix $\Delta \mathbf{W}$, constructed from the product of two smaller trainable matrices: $\mathbf{A} \in \mathbb{R}^{h \times r}$ and $\mathbf{B} \in \mathbb{R}^{r \times h}$, where $h$ is the hidden dimension and $r \ll h$ is a rank parameter that controls adaptation capacity. This results in a new weight formulation:

$$\mathbf{W}_{new} = \mathbf{W}_{pretrained} + \mathbf{AB}.$$

By keeping the original weights $\mathbf{W}_{pretrained}$ frozen and only training the low-rank components, LoRA enables task-specific adaptation—in our case, alignment to multilingual guardrail behavior—while maintaining the model's general language understanding.

In building SEALGUARD, we incorporate LoRA into key components of the SeaLLM model: the word embedding layer, the self-attention blocks, and the feed-forward neural networks (FFNN).

### 3.4 Detect and Block Multilingual Unsafe Prompts

In Step ⑤, SEALGUARD intercepts user prompts before they reach an LLM-powered system, identifying and blocking unsafe inputs. Similar to LlamaGuard [16], SEALGUARD frames unsafe prompt detection as sequence generation. Given the final transformer decoder output $\mathbf{H} \in \mathbb{R}^{l \times h}$—with $l$ as the sequence length and $h$ as the hidden size—a linear layer projects each hidden state $\mathbf{H}i$ to a vocabulary distribution using a weight matrix $\mathbf{W}_{lm} \in \mathbb{R}^{h \times v}$ and bias $\mathbf{b}_{lm} \in \mathbb{R}^v$, where $v$ = 152,064. This transformation produces a vector of raw prediction scores, referred to as logits, for each token position.

These logits are then passed through a softmax function to compute probabilities over all possible vocabulary tokens. Guided by the chat template introduced in Step ①, the model generates a response sequence, where the first token explicitly states whether the input is "safe" or "unsafe". We use greedy decoding to generate tokens, selecting at each step the one with the highest probability:

$$t_i = \arg\max_k \text{softmax}(\mathbf{H}_i \mathbf{W}_{lm} + \mathbf{b}_{lm})_k$$

where $k$ indexes the vocabulary. Generation stops at either the special end-of-text token "<|im_end|>" or a predefined length limit. The first generated token is used as the model's safety decision.

## 4 EXPERIMENTAL DESIGN

In this section, we outline the motivation behind our four research questions, introduce the proposed benchmark dataset that serves as our experimental dataset, describe the types of unsafe prompts and jailbreak attacks studied, present the baseline guardrails, and detail our experimental setup.

### 4.1 Research Questions

To evaluate the effectiveness of multilingual guardrails for Southeast Asian languages, we formulate the following four research questions:

**RQ1) What is the impact of multilingual unsafe prompts in Southeast Asian languages on the performance of existing guardrails?** LlamaGuard [16] achieves state-of-the-art performance on English data, reaching 95% accuracy as a runtime guardrail for classifying safe and unsafe prompts. However, its training data is primarily in English. As a result, a key limitation arises: Inan *et al.* Inan et al. [16] speculated that LlamaGuard may be vulnerable to prompts written in other languages. This highlights a research gap in multilingual safety alignment, where existing guardrails may fail to generalize across diverse linguistic inputs. Yet, little is known about how multilingual unsafe prompts affect the performance of these AI guardrails. Thus, we investigate the impact of multilingual unsafe prompts on current AI guardrail systems.

**RQ2) What is the impact of multilingual jailbreak prompts in Southeast Asian languages on the performance of existing guardrails?** While general multilingual unsafe prompts already challenge existing guardrails, multilingual jailbreak prompts pose an even greater risk. These prompts are intentionally crafted to bypass safety mechanisms, circumventing the protections that guardrails provide. Yet, current studies have not examined how such jailbreak prompts affect the performance of AI guardrail systems in multilingual settings. Thus, we investigate the impact of multilingual jailbreak prompts on existing AI guardrails.

**RQ3) How effective is our SEALGuard at defending against multilingual unsafe and jailbreak prompts in Southeast Asian languages?** To address the gap in multilingual safety alignment, we proposed a multilingual guardrail capable of detecting unsafe inputs across diverse linguistic settings. In Section 3, we describe how we adapt a multilingual language model into a guardrail using low-rank adaptation (LoRA). However, the effectiveness of this approach remains unknown. Thus, we evaluate the performance of our proposed SEALGuard against existing baseline guardrails, focusing on its ability to detect both multilingual unsafe and jailbreak prompts while minimizing false positives.

**RQ4) What are the contributions of adaptation strategies and model size to the performance of our SEALGuard multilingual guardrail?** Our SEALGuard multilingual guardrail relies on adapting a general-purpose multilingual language model into a safety guardrail using low-rank adaptation (LoRA). While LoRA serves as the primary adaptation strategy, the impact of different adaptation approaches and model sizes on guardrail performance remains unclear. To better understand these factors, we formulate this research question to conduct an ablation study, examining how adaptation strategies and model size influence the effectiveness of SEALGuard.

### 4.2 SEALSBench: A Multilingual Safety Alignment Benchmark in Southeast Asian Languages

To address the four research questions, we construct a multilingual LLM safety benchmark, SEALSBench, comprising 18,846 safe prompts, 8,959 unsafe prompts, and 1,799 jailbreak prompts (unsafe), totaling 29,604 prompts. Jailbreak prompts are a specific category of unsafe inputs designed to bypass or mislead LLM safety mechanisms. Figure 3 presents the distribution of safe and unsafe prompts, along with a detailed breakdown of unsafe categories and jailbreak prompt types. These prompts were initially written in English and then translated into nine Southeast Asian languages to assess the effectiveness of multilingual guardrails across diverse linguistic contexts. Figure 4 summarizes the data construction workflow. Below, we outline the step-by-step process used to construct this dataset.

**Step 1: Data Collection**. We extracted 8,959 unsafe prompts from the BeaverTails dataset by Ji *et al.* [18], covering ten categories of unsafe prompts introduced in Section 4.3. These categories capture a broad range of safety concerns related to input prompts submitted to LLMs, consistent with prior LLM safety studies [11, 31, 33]. Moreover, we extracted 1,799 unsafe prompts from four additional benchmark datasets: Do-Not-Answer [42], CatQA [4], AdvBench [52], and Forbidden Questions [36] and transformed them into jailbreak prompts. These additional datasets help avoid data contamination and ensure our jailbreak attacks have different distributions from the unsafe prompts in the BeaverTails dataset. We then transformed them into jailbreak prompts using nine jailbreak attack strategies introduced in Section 4.4. This allows us to cover a special class of unsafe prompts—jailbreak prompts—that are designed to bypass safety guardrails and reflect jailbreak attempts that may occur in real-world use. To evaluate whether safe inputs will be incorrectly blocked by SEALGuard, we collect 18,846 safe prompts from the Alpaca dataset by Taori *et al.* [39]. These prompts represent safe interactions with LLMs that guardrails should allow without blocking. In summary, our dataset contains a total of 29,604 prompts, including 8,959 unsafe prompts, 1,799 jailbreak prompts, and 18,846 safe prompts.

**Step 2: Multilingual Translation into Southeast Asian Languages.** The collected 29,604 prompts were originally written in English. In this study, we focus on nine Southeast Asian languages supported by SeaLLM [29], covering major linguistic families in the region: Chinese (Zho), Indonesian (Ind), Vietnamese (Vie), Thai (Tha), Khmer (Khm), Lao, Malay (Msa), Burmese (Mya), and Tagalog (Tgl). To support multilingual evaluation, we used the Google Translate API [13] to translate all prompts into these nine languages. This results in a total of 296,040 prompts, including the original English and its translations into nine Southeast Asian languages. These collectively form our multilingual safety benchmark dataset, SouthEast Asian Languages Safety Benchmark, SEALSBench.

### 4.3 Studied Unsafe Categories

Our SEALSBench dataset consists of ten unsafe categories to support the evaluation of LLM safety alignment. Each category is labeled as **C1** through **C10**, where **C** stands for Category: C1: Violent Criminal Activity (1,729 samples), C2 – Non-Violent Criminal Conduct (4,196 samples), C3 – Child Sexual Abuse (145 samples), C4 – False and Defamatory Claims (1,415 samples), C5 – Hazardous Professional Guidance (362 samples), C6 – Personal Information Exposure (852 samples), C7 – Discriminatory and Hateful Expression (1,022 samples), C8 – Self-Destructive Behavior Promotion (87 samples), C9 – Explicit Sexual Material (342 samples), C10 – Misinformation & Extremist Content (807 samples).

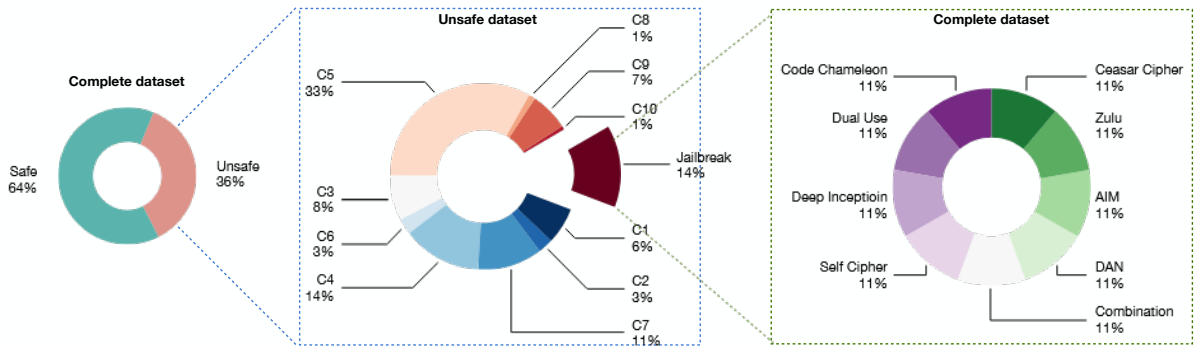

Figure 3: The distribution of safe and unsafe prompts, along with category-wise breakdowns of unsafe and jailbreak prompts.

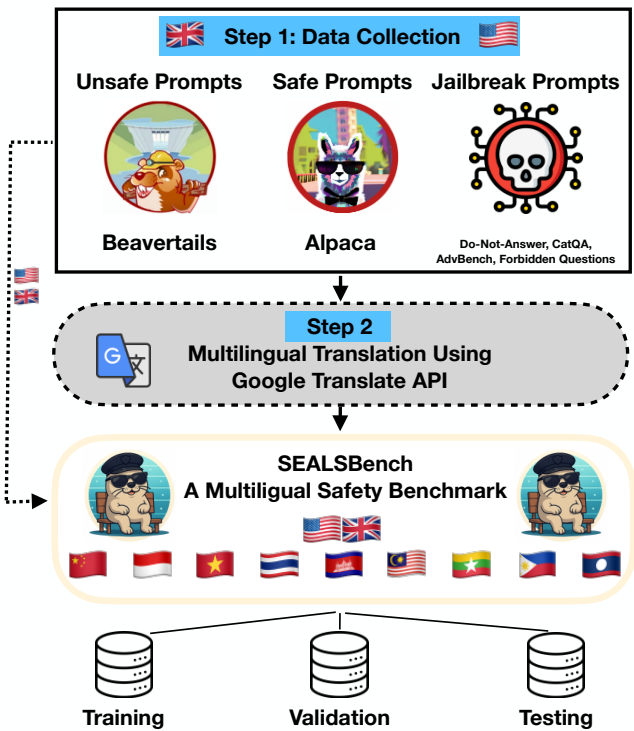

Figure 4: Workflow for Constructing our SEALS-Bench Dataset.

### 4.4 Studied Jailbreak Attacks

We consider nine different jailbreak attacks spanning these three major categories.

**Obfuscation-Based Attacks.** These techniques aim to circumvent the safety mechanisms of large language models (LLMs) by disguising unsafe prompts through various forms of obfuscation.

**Caesar Cipher [48]** employs systematic character replacement techniques, where responses requested in matching encrypted formats can potentially bypass both input and output safety mechanisms.

**Zulu [47]** exploits low-resource language vulnerabilities by reformulating harmful prompts in less-supported languages, then manipulating the model to translate them back into unsafe content, thereby evading guardrail systems.

**Template-Based Attacks.** These attacks utilize pre-constructed frameworks or prompt structures that exploit predictable patterns and known vulnerabilities within LLMs.

**AIM [17]** manipulates the model into adopting a predefined persona that operates outside normal ethical constraints, promoting unethical, illegal, and harmful responses through character role-play.

**DAN [37]** creates scenarios where the model 'believes' normal restrictions no longer apply, often framing interactions as role-playing exercises to bypass content filters.

**Combination (Prefix injection + Refusal Suppression) [43]** combines multiple techniques, including prefix injection (using innocuous openings) with refusal suppression instructions, constraining the model's ability to generate standard refusal responses.

**Self Cipher [49]** prompts the model to act as an expert in undefined cipher systems, leveraging the model's internal encoding capabilities to implicitly encrypt queries and decrypt outputs without explicit encoding rules.

**Deep Inception [22]** employs multi-layered narrative structures that progressively guide the model toward restricted behavior through incremental logical steps, often embedded within fictional scenarios.

**Code-Based Attacks.** These attacks exploit LLMs' programming capabilities by disguising harmful content within technical instructions or programming logic.

**Dual use [19]** combines code injection techniques with payload splitting to craft harmful prompts that appear as legitimate programming tasks while containing malicious intent.

**Code Chameleon [24]** reformulates harmful instructions as code completion tasks, using custom encryption functions embedded within programming contexts to enable decryption and execution of harmful queries while bypassing safety mechanisms.

### 4.5 Baseline Guardrails

(1) **LlamaGuard [16]:** An LLM-based guardrail fine-tuned on a proprietary moderation dataset to classify prompts as "safe" or "unsafe." We evaluate two variants: "Llama-Guard-3-8B" and the smaller "Llama-Guard-3-1B".
(2) **OpenAI Moderation [25]:** A GPT-based moderation system trained via active learning on public data. It flags prompts as unsafe if the returned boolean is "True", covering diverse safety categories.

### 4.6 Experimental Setup

**Data Splitting.** We split our dataset into 5% training (14,800 samples), 5% validation (14,800 samples), and 90% testing (266,444 samples). We ensure that all language variants of a given prompt remain in the same split by partitioning based on unique English prompt IDs rather than random sampling.

**Model Implementation and Optimization.** We developed SEAL-GUARD using Transformers [44] and PyTorch [32], fine-tuning the multilingual SeaLLMs-v3-7B-Chat [50] with LoRA [15]. Training was conducted on an AMD 5950X CPU with two NVIDIA RTX 3090 GPUs. Input prompts were wrapped in a chat template, and the model was trained to autoregressively generate the prompt followed by a classification token ("safe" or "unsafe"). We optimized the model using Cross-Entropy Loss, masking prompt tokens during loss computation to focus learning on the classification output. The loss is defined as

$$\mathcal{L}_{CE} = -\sum_{t=1}^{T} \log P_\theta(y_t \mid y_{<t}, x) \quad (1)$$

where $T$ is the length of the target sequence, $y_t$ is the target token at position $t$, $y_{<t}$ are previously generated tokens, $x$ is the input, and $P_\theta$ is the model's predicted probability distribution parameterized by $\theta$.

**Hyper-Parameter Settings.** We followed standard hyperparameter settings commonly used for LoRA fine-tuning. Specifically, we used a learning rate of 1e-4 with a LoRA dropout rate of 0.05. The rank ($r$) of the LoRA modules was set to 8, and the scaling factor ($\alpha$) was set to 32. We applied gradient clipping with a maximum gradient norm of 1.0. For learning rate scheduling, we used a cosine scheduler with 5% of the total training steps allocated to warm-up. The complete training recipe of our SEALGUARD approach is available in our replication package at https://github.com/awsm-research/SEALGuard.

## 5 EXPERIMENTAL RESULTS

In this section, we present the results for our three research questions.

### (RQ1) What is the impact of multilingual unsafe prompts in Southeast Asian languages on the performance of existing guardrails?

**Approach.** To assess the impact of multilingual unsafe prompts in Southeast Asian languages on the performance of existing guardrails, we use 80,601 unsafe prompts from the SEALSBENCH testing set, written in English and nine Southeast Asian (SEA) languages. We first evaluate guardrail performance on the English prompts to establish baselines in their familiar language. We then assess performance on the non-English SEA prompts to measure the impact of multilingual inputs. Specifically, we evaluate two state-of-the-art language model-based guardrails: LlamaGuard [16] and OpenAI Moderation [25]. We use Defense Success Rate (DSR) as our primary metric, which evaluates guardrail capability to defend against unsafe prompts:

- **Defense Success Rate (DSR):**

$$\text{DSR} = \frac{TP}{TP + FN}$$

where $TP$ is the number of true positives (correctly detected unsafe/jailbreak prompts) and $FN$ is the number of false negatives (missed unsafe/jailbreak prompts).

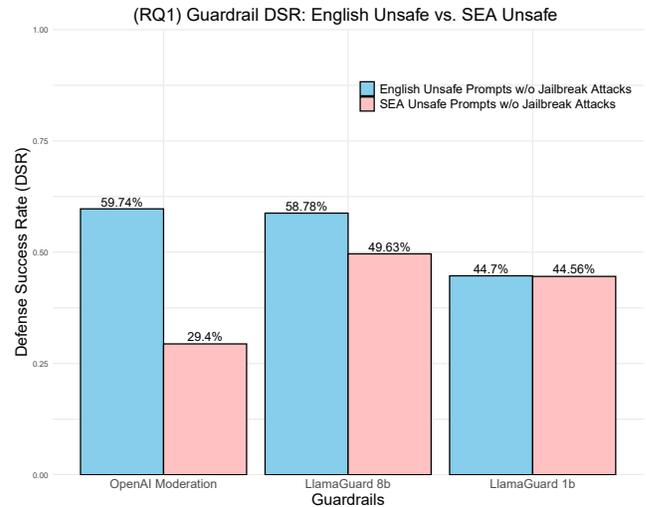

**Figure 5: DSR result of unsafe prompts in English and SEA languages.**

**Results.** Figure 5 presents the defense success rate (DSR) of the three baseline guardrails, comparing their performance across two scenarios: unsafe prompts in English (blue bars) versus multilingual unsafe prompts, excluding jailbreak prompts, in Southeast Asian languages (red bars).

**LlamaGuard 8B's DSR substantially decreases by 9%, declining from 58.78% to 49.63% when defending against multilingual unsafe prompts.** Similarly, OpenAI Moderation's DSR experiences a substantial decline of 30%, dropping from 59.74% to 29.4% when encountering multilingual unsafe prompts. On the other hand, LlamaGuard 1B achieves a lower performance of 45% for both English unsafe prompts and multilingual unsafe prompts than

LlamaGuard 8B. **These findings reveal that the effectiveness of state-of-the-art guardrails is decreased when defending multilingual unsafe prompts, highlighting the need for multilingual guardrails capable of effectively defending such multilingual prompts.**

## (RQ2) What is the impact of multilingual jailbreak prompts in Southeast Asian languages on the performance of existing guardrails?

**Approach**. To assess the impact of multilingual jailbreak prompts in Southeast Asian languages on guardrail performance, we use 8,060 unsafe prompts in English, and 16410 jailbreak prompts in English and nine SEA languages from our SEALSBench test set. We first evaluate guardrail performance on English unsafe prompts, which yields the same baselines as in RQ1. We then assess performance on multilingual jailbreak prompts to measure the effect of multilingual jailbreak inputs, using the same baseline and metric (DSR) as in RQ1.

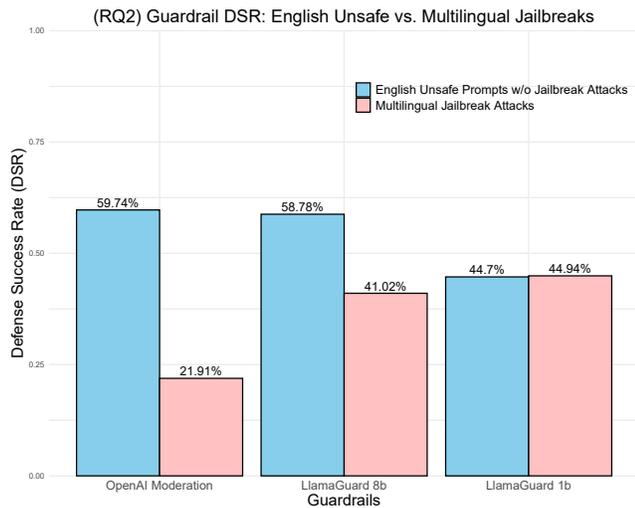

**Figure 6: DSR result of unsafe prompts in English and Multilingual jailbreak prompts.**

**Results**. Figure 6 presents the defense success rate (DSR) of the three baseline guardrails, comparing their performance across two scenarios: unsafe prompts written in English without jailbreak attacks (blue bars) versus multilingual jailbreak attacks in Southeast Asian languages and English (red bars).

**LlamaGuard 8B's DSR substantially decreases by 18%, declining from 58.78% to 41.02% when defending against multilingual jailbreak prompts.** Similarly, OpenAI Moderation's DSR experiences a substantial decline of 38%, dropping from 59.74% to 21.91% when encountering multilingual jailbreak prompts. As in RQ1, LlamaGuard 1B still achieves a lower performance of 45% for both English unsafe prompts and multilingual jailbreak prompts than LlamaGuard 8B. **These findings highlight that the effectiveness of state-of-the-art guardrails is decreased when defending multilingual jailbreak prompts, highlighting the need for multilingual jailbreak-aware guardrails capable of effectively defending such multilingual jailbreak prompts.**

## (RQ3) How effective is our SEALGuard at defending against multilingual unsafe and jailbreak prompts in Southeast Asian languages?

**Approach**. To address this RQ, we compare our SEALGuard approach with LlamaGuard and OpenAI Moderation using the full multilingual testing set from our SEALSBench dataset. This set includes 169,433 safe, 80,601 unsafe, and 16,410 jailbreak prompts written in English and nine Southeast Asian languages. We use the same Defense Success Rate (DSR) as in RQ1 and RQ2 to evaluate guardrail effectiveness in defending against unsafe and jailbreak prompts. Beyond defense capability, maintaining a low false alarm rate is also critical to avoid misclassifying safe content. Thus, we use precision to measure the accuracy of the guardrail in avoiding false alarms—i.e., how well it identifies only truly unsafe prompts without misclassifying safe ones. We also use the F1-Score to capture the overall balance between correctly detecting unsafe prompts and minimizing false positives. Formally:

- **Precision**:

$$\text{Precision} = \frac{TP}{TP + FP}$$

where $TP$ is the number of true positives (correctly detected unsafe/jailbreak prompts) and $FP$ is the number of false positives (safe prompts incorrectly classified as unsafe).
- **F1-Score**:

$$F1 = 2 \cdot \frac{\text{Precision} \cdot \text{DSR}}{\text{Precision} + \text{DSR}}$$

**Results**. Figure 7 presents the experimental results of our SEALGuard and the three baseline approaches according to our three evaluation measures (i.e., DSR, Precision, and F1-Score).

**Our SEALGuard achieves an F1-Score of 98%, which is 58%, 52%, and 34% better than the LlamaGuard-3-1B, OpenAI Moderation, and LlamaGuard-3-8B respectively.** In terms of DSR, Figure 7 shows that our SEALGuard achieves the highest DSR of 97%, while the three baselines achieve 49%, 45%, and 31%, respectively. This finding indicates that SEALGuard substantially improves the DSR by 48%, 52%, and 66%. In terms of precision, Figure 7 shows that SEALGuard achieves the highest precision of 99%, while the three baselines achieve 96%, 91%, and 36%, respectively. This finding demonstrates that SEALGuard substantially improves precision by 3%, 8%, and 63%.

In summary, these findings confirm that our SEALGuard approach is effective in defending against multilingual unsafe and jailbreak prompts while maintaining a low false positive rate. **These results demonstrate that SEALGuard effectively overcomes a key limitation of the state-of-the-art guardrail, LlamaGuard—namely, its reduced effectiveness in defending against multilingual unsafe and jailbreak prompts, as shown in RQ1 and RQ2.**

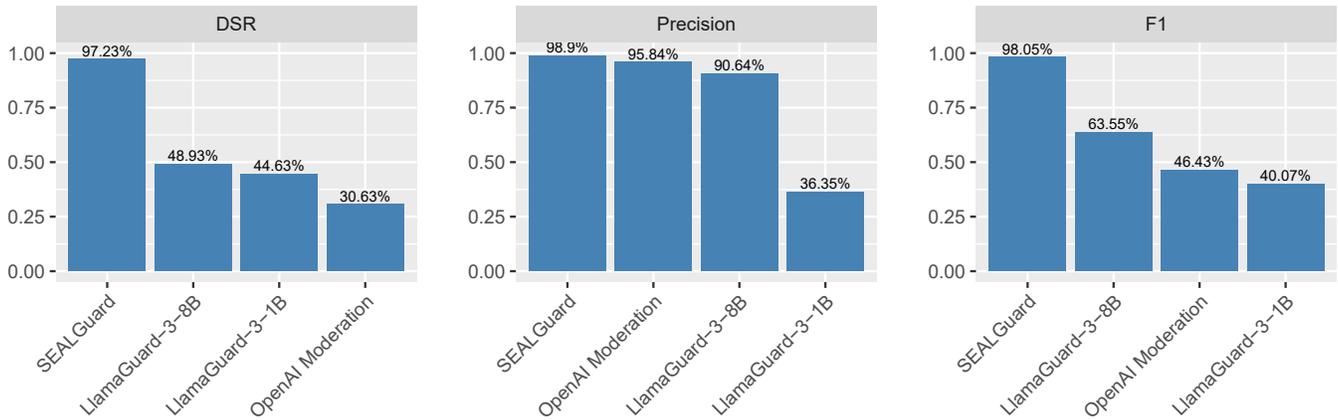

Figure 7: (RQ3) The experimental results of our SEALGuard and the three baseline comparisons classifying safe and unsafe (including jailbreaks) prompts. (↗) Higher F1, DSR, Precision, Accuracy = Better.

Table 1: (RQ4 Results) The performance comparisons of our SEALGuard approach and the five variants to analyze the contributions of the adaptation strategy and model size.

| Methods | DSR | Precision | F1-Score |
|---|---|---|---|
| SeaLLM-V3-7B-Chat + LoRA (SEALGuard) | **97.23** | **98.90** | **98.05** |
| SeaLLM-V3-1.5B-Chat + LoRA | 96.5 | 98.20 | 97.34 |
| SeaLLM-V3-7B-Chat + NeMo | 49.88 | 77.17 | 60.59 |
| SeaLLM-V3-1.5B-Chat + NeMo | 89.45 | 34.89 | 50.20 |
| SeaLLM-V3-7B-Chat | 15.07 | 89.51 | 25.80 |
| SeaLLM-V3-1.5B-Chat | 18.48 | 60.14 | 28.27 |

## (RQ4) What are the contributions of adaptation strategies and model size to the performance of our SEALGuard multilingual guardrail?

**Approach.** To address this RQ, we conduct an ablation study to evaluate the contribution of each component in SEALGuard. Our approach combines two key elements: the SeaLLM-v3-7B-Chat model (7B parameters) and a LoRA-based adaptation strategy that aligns a general-purpose language model with guardrail objectives. To assess the effectiveness of the adaptation, we compare our method with two variants: (1) applying the chat template introduced in Section 3.1 directly to SeaLLM-v3-7B-Chat without adaptation, and (2) using the NVIDIA NeMo toolkit with the same model, an off-the-shelf guardrail framework that requires no fine-tuning. To examine the impact of model size, we also compare against a smaller model, SeaLLM-v3-1.5B-Chat. In summary, we evaluate six models in this RQ:

- SeaLLM-V3-7B-Chat + LoRA (SEALGuard): our proposed approach by applying LoRA on a multilingual language model.
- SeaLLM-V3-1.5B-Chat + LoRA: applying LoRA on a smaller multilingual language model to study the impact of model size on performance.
- SeaLLM-V3-7B-Chat + NeMo: applying the NeMo framework for adaptation to study the effect of alternative adaptation strategies.
- SeaLLM-V3-1.5B-Chat + NeMo: applying the NeMo framework on a smaller model to analyze both adaptation strategy and model size impact.
- SeaLLM-V3-7B-Chat: no adaptation, used to isolate and evaluate the contribution of adaptation strategies.
- SeaLLM-V3-1.5B-Chat: no adaptation, used to evaluate the baseline performance of a smaller model without any adaptation.

**Results.** Table 1 presents the performance comparison of our SEALGuard approach and five variants to analyze the contributions of adaptation strategies and model size.

**LoRA adaptation is the key component driving the effectiveness of our SEALGuard approach.** Within SEALGuard, the LoRA module alone contributes 72% of the total F1-Score. Specifically, when comparing "SeaLLM-V3-7B-Chat + LoRA (SEALGuard)" with "SeaLLM-V3-7B-Chat" (without LoRA), the F1-Score drops from 98% to 26%, highlighting a 72% contribution by LoRA.

In addition, comparing "SeaLLM-V3-7B-Chat + NeMo" with "SeaLLM-V3-7B-Chat + LoRA (SEALGuard)" shows an F1-Score increase from 61% to 98%, representing a 37% improvement attributed to LoRA. Similarly, for the smaller variant of SEALGuard, "SeaLLM-V3-1.5B-Chat + LoRA" outperforms both "SeaLLM-V3-1.5B-Chat" and "SeaLLM-V3-7B-Chat + NeMo" by 70% and 38% in F1-Score, respectively. **These results demonstrate that LoRA is the most effective adaptation strategy for aligning a general-purpose language model with a guardrail.** In summary, these findings validate the design rationale of SEALGuard, showing that LoRA adaptation is the primary driver of performance gains, while model size plays a comparatively minor role—even models with around 1B parameters can achieve promising results with our approach.

## 6 DISCUSSION

Our experimental results confirm the effectiveness of SEALGuard in defending against multilingual unsafe and jailbreak prompts, showing substantial improvements over existing guardrails. However, its performance across different languages, unsafe prompt types, and jailbreak prompts remains unclear. Thus, in this section, we analyze SEALGuard 's performance across these three dimensions.

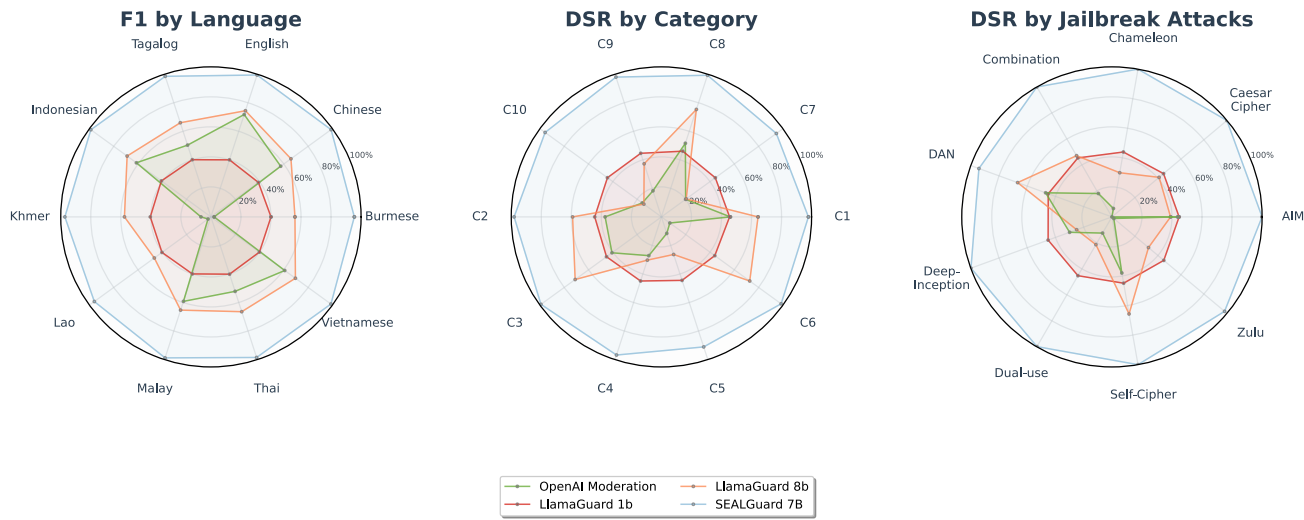

Figure 8: (Discussion) The experimental results for our SEALGuard and the other five baseline models classifying in category, language, and jailbreak attacks.

## 6.1 Performance Across Different Languages

To assess the cross-lingual robustness of our SEALGuard approach, we analyze the performance of our SEALGuard across ten different languages. Specifically, the evaluation includes 80,601 unsafe prompts, 16,410 jailbreak prompts, and 169,433 safe prompts from our SEALSBench dataset. We focus on the F1-score as it captures both the model's ability to defend against unsafe prompts and its tendency to raise false alarms on safe ones.

The left part of Figure 8 shows the F1-scores of our SEAL-Guard across the ten studied languages. **SEALGuard consistently achieves the highest performance in all languages, effectively mitigating the language-specific vulnerabilities present in existing guardrails.** It also outperforms the strongest baseline, LlamaGuard 8B, by a clear margin, as illustrated in Figure 8. This analysis confirms the cross-lingual robustness of our SEALGuard approach.

## 6.2 Performance on Different Unsafe categories

To assess the robustness of SEALGuard across different types of unsafe content, we analyze its performance on ten different unsafe categories. As introduced in 4.3, we used 10 unsafe categories labeled as C1 to C10 in our test dataset, totaling 80,601 prompts. We adopt DSR as the evaluation measure because this analysis involves only unsafe prompts, allowing DSR to assess single-class performance without interference from false positives associated with safe prompts.

The middle part of Figure 8 presents the DSR of SEALGuard compared to baseline guardrails across ten unsafe categories. LlamaGuard-8B exhibits low Defense Success Rate (DSR below 30%) in several categories, including C5 (Hazardous Professional Guidance), C7 (Discriminatory and Hateful Expression), and C10 (Misinformation & Extremist Content). In contrast, **SEALGuard demonstrates consistently high performance across all categories, with DSR exceeding 95% in every category, confirming its robustness across diverse types of unsafe content.**

## 6.3 Performance on Different Jailbreak Prompts

To evaluate the robustness of SEALGuard across different jailbreak categories, we analyze its performance against nine jailbreak attacks. As described in Section 4.4, this analysis covers 10 jailbreak categories from our test dataset, totaling 16,410 prompts. We use Defense Success Rate (DSR) as the evaluation metric, as the analysis focuses solely on jailbreak prompts without any safe examples, making DSR the appropriate measure for assessing single-class performance.

The right part of Figure 8 presents the DSR of SEALGuard across nine jailbreak categories. While LlamaGuard-8B shows low Defense Success Rate (DSR below 30%) against Deep-Inception, Chameleon, Zulu, and Dual-use jailbreak prompts, **SEALGuard achieves consistently high performance across all jailbreak types, with DSR exceeding 95% in every category, confirming its robustness against diverse jailbreak attacks.**

## 7 THREATS TO VALIDITY

*Threats to internal validity* relate to factors within our study that may affect the accuracy of the findings. One such threat is the potential variation in translation accuracy when converting English prompts into Southeast Asian languages, which may affect guardrails' detection performance. To mitigate this, we rely on consistent use of the Google Translate API and make our translation process publicly available to promote transparency and reproducibility. Another internal threat is the limited diversity in unsafe and jailbreak prompts, which could bias evaluation results. To mitigate this, we curate SEALSBench from six diverse data sources [4, 18, 36, 39, 42, 52], including ten unsafe categories and nine jailbreak categories, thereby enhancing the representativeness of the dataset.

*Threats to external validity* relate to the degree to which our findings can be generalized to other LLM safety alignment datasets used for evaluating guardrail performance. While our SEALGuard approach is assessed using our curated SEALSBench dataset, the results may not necessarily generalize to other datasets. To mitigate this, we incorporate prompts from six diverse sources [4, 18, 36, 39, 42, 52] during the data collection step when constructing our SEALSBench dataset, ensuring a broad representation of safe, unsafe, and jailbreak prompts. The final dataset comprises over 260,000 prompts written in ten different languages.

## 8 CONCLUSION

In this paper, we empirically show that LlamaGuard's Defense Success Rate (DSR) drops by 9% and 18% under multilingual unsafe and jailbreak prompts, revealing a critical gap in multilingual safety alignment for LLM software systems. To address this, we introduce SEALGuard, a multilingual guardrail adapted via LoRA from a general-purpose multilingual model, focusing on Southeast Asian languages. We also present SEALSBench, a benchmark dataset of over 260K prompts across English and nine Southeast Asian languages, covering safe, unsafe, and jailbreak scenarios. SEALGuardachieves 97% DSR, 99% precision, and 98% F1-score, outperforming LlamaGuard by 48%, 8%, and 34%, respectively, while maintaining robust performance across languages and unsafe types.